\documentclass[letterpaper]{article} 
\usepackage{aaai23}
\usepackage{times}  
\usepackage{helvet}  
\usepackage{courier}  
\usepackage[hyphens]{url}  
\usepackage{graphicx} 
\usepackage{natbib}  
\usepackage{caption} 
\usepackage{algorithm}
\usepackage{algorithmic}

\usepackage{amsmath}
\usepackage{amssymb}
\usepackage{booktabs}
\usepackage{comment}
\usepackage{multicol}
\usepackage{multirow}
\usepackage{enumitem}
\usepackage{rotating}
\usepackage{todonotes}
\usepackage{subcaption}
\usepackage{newfloat}
\usepackage{listings}
\DeclareCaptionStyle{ruled}{labelfont=normalfont,labelsep=colon,strut=off} 
\floatstyle{ruled}
\newfloat{listing}{tb}{lst}{}
\floatname{listing}{Listing}
\title{Towards Human-Interpretable Prototypes for Visual Assessment of Image Classification Models}

\author {
    Poulami Sinhamahapatra,\textsuperscript{\rm 1}
    Lena Heidemann, \textsuperscript{\rm 1}
    Maureen Monnet, \textsuperscript{\rm 1}
    Karsten Roscher \textsuperscript{\rm 1}
}
\affiliations {
    \textsuperscript{\rm 1}  Fraunhofer IKS
}

\usepackage{bibentry}

\begin{document}

\maketitle

\begin{abstract}
  Explaining black-box Artificial Intelligence (AI) models is a cornerstone for trustworthy AI and a prerequisite for its use in safety critical applications such that AI models can reliably assist humans in critical decisions. However, instead of trying to explain our models post-hoc, we need models which are interpretable-by-design built on a reasoning process similar to humans that exploits meaningful high-level concepts such as shapes, texture or object parts. Learning such concepts is often hindered by its need for explicit specification and annotation up front. Instead, prototype-based learning approaches such as ProtoPNet claim to discover visually meaningful prototypes in an unsupervised way. In this work, we propose a set of properties that those prototypes have to fulfill to enable human analysis, e.g. as part of a reliable model assessment case, and analyse such existing methods in the light of these properties. Given a \textit{`Guess who?'} game, we find that these prototypes still have a long way ahead towards definite explanations. We quantitatively validate our findings by conducting a user study indicating that many of the learnt prototypes are not considered useful towards human understanding. We discuss about the missing links in the existing methods and present a potential real-world application motivating the need to progress towards truly human-interpretable prototypes.
\end{abstract}


\section{Introduction}

\label{sec:intro}

In recent years, Deep Neural Networks (DNNs) have been shown to be increasingly proficient in solving more and more complex tasks. Increasing complexity of tasks lead DNNs to churn billions of parameters, vast pools of unstructured data and non-human understandable internal representations to arrive at these spectacular results. However, this only adds to the complexity and opacity of the already \textit{black-box} DNNs. This calls for an increasing need to promote eXplainable Artificial Intelligence (XAI) approaches, for improving interpretability, transparency, and trustworthiness of AI ~\cite{adadiPeekingBlackBoxSurvey2018,barredoarrietaExplainableArtificialIntelligence2020}. 
It is even more critical to reason and explain the decision-making process of DNNs, when such decisions are applied in safety-critical use-cases like self-driving cars or medical diagnosis \cite{tjoaSurveyExplainableArtificial2020}. These cases not only demand higher accountability to figure out why and what, if things go wrong but also a provision to assess, debug and audit in cases of failure modes. 

In general, XAI approaches try to map the internal learnt representations of DNNs into human-interpretable formats. However, what constitutes a \textit{sufficiently human-understandable interpretation} is still largely subjective of the XAI approach itself, whether it is post-hoc or inherently interpretable or seeks local or global explanations and so on. One novel direction is associated with learning representations which can be explicitly tied to higher level concepts relevant to humans, e.g. predicting an image of a bird \textit{red-billed hornbill} could depend on the presence of concepts like \textit{red bill}.   
 However, this requires explicit prior knowledge of relevant concepts or attributes and relies on correspondingly annotated datasets. Instead, \textit{unsupervised discovery} of the relevant parts or prototypes would enable the use of existing large-scale datasets and open such approaches to more diverse use-cases. Such prototypes can represent distinct human-understandable concepts or sub-parts, e.g. \textit{beaks, wings, tails} which could together predict a \textit{bird}. While learning such representations in an unsupervised scenario (absence of concept-level annotation) in itself is a challenge, the other difficulty lies in visualising such implicitly learnt representations understandable to human eye. 

A very prominent line of work based on learning interpretable prototypes has emerged where the focus is to learn representative parts for the downstream task by finding the closest sample to a given learnt representation as \textit{prototypes} ~\cite{liDeepLearningCaseBased2018,chenThisLooksThat2019a,xuAttributePrototypeNetwork2020,vanlooverenInterpretableCounterfactualExplanations2021,nautaThisLooksThat2021a,nautaNeuralPrototypeTrees2021}. 
How useful are these interpretations with respect to human assessment of the model's inner workings and potential insufficiencies? 

In this work, we closely investigate the performance of selected interpretable prototype-based approaches in terms of qualitative interpretation using a network called  \textit{Prototypical Parts Network (ProtoPNet)} ~\cite{chenThisLooksThat2019a} and subsequent variants ~\cite{nautaNeuralPrototypeTrees2021,gautamThisLooksMore2021}. 
They are designed to learn representations of certain parts of the training image class (\textit{prototypes}) and then find the (parts of) test samples similar to the prototypes  (``this-looks-like-that'') based on similarity scores. To this end, our key contributions are: 
\begin{enumerate}
    \item We design common setup of experiments and accordingly propose requisite properties (\textit{Desiderata}, Section \ref{sec:desiderata}) towards learning interpretable prototypes beneficial for the human assessment of a model, 
    \item We analyse existing methods with respect to these properties for both real-world and synthetic datasets (Section \ref{sec:results-real}, \ref{sec:results-synthetic}),
    \item We provide quantitative results based on a user study to validate our findings on real images (Section \ref{sec:results-userstudy}) and 
    \item Finally, we motivate the application of interpretable prototypes using real-world out-of-distribution (OOD) detection task (Section \ref{sec:results-ood}) and conclude with imminent challenges and potential directions (Section \ref{sec:conclusion}). 
\end{enumerate}


\section{Interpretability So Far}
\label{sec:2}

According to \cite{paezPragmaticTurnExplainable2019}, \textit{Interpretability} means an AI system's decision can be explained globally or locally and the system's purpose as well as decision can be understood by a human actor. 
There exists a vast pool of XAI literature  \cite{schwalbeComprehensiveTaxonomyExplainable2021,holzingerExplainableAIMethods2022}, also pertaining to visual tasks \cite{nguyenUnderstandingNeuralNetworks2019,samekExplainableArtificialIntelligence2019}.
Following are some major distinguishing factors towards choosing a given XAI approach:

\noindent \textbf{Local vs. Global Explanation} \par
Several vision approaches using DNNs focus on local explanations \cite{zhangVisualInterpretabilityDeep2018, ribeiro2016should} 
limited to specific few samples or highlighting specific parts of the image that the DNNs attended to for a given decision (i.e. regions with highest \textit{attribution}), say by generating heatmaps. These methods often involve generating saliency-based activation maps \cite{zhouLearningDeepFeatures2016}, local sensitivity based on gradients \cite{sundararajanAxiomaticAttributionDeep2017}, or using perturbations \cite{fongInterpretableExplanationsBlack2017} or relevance back-propagation \cite{shrikumar17a} such as LRP (Layer-wise Relevance Propagation) \cite{bachPixelWiseExplanationsNonLinear2015}. However, all these approaches are often local analysis. In contrast, global explanations provide analysis on the models as a whole, independent of individual examples \cite{lundberg2020local}, e.g. mapping certain concepts to internal latent representations. This provides a wider scope for general applicability. For some tasks, like OOD detection, global explanations are even necessary to detect new OOD samples. 


\noindent \textbf{Post-Hoc vs. Interpretable-by-Design Explanation}\par

Most of the above local explanation methods
are also post-hoc interpretations, which involve taking a pre-trained model and then identifying relevant features  via attribution  or trying to understand the inner workings a posteriori. Since these explanations are not tied to the inner workings of the model, they can be unreliable. Thus, we need inherently interpretable models, i.e. interpretable-by-design (IBD), such that DNNs are designed in a way to make internal representations interpretable. IBD methods have gained much momentum because if we want our models to be explainable, we need to consciously design them to be interpretable \cite{rudinStopExplainingBlack2019}. One recent direction towards IBD models is to map human-understandable concepts or prototypes into internal representations, e.g. by embedding an interpretable layer into the network like in concept bottleneck models (CBM) \cite{kohConceptBottleneckModels2020a}, ProtoPNet models etc., or by enforcing single concepts into a model by including their outputs in the loss function of the model \cite{zhangInterpretableConvolutionalNeural2018}.


\noindent \textbf{Explicitly Specified vs. Implicitly Derived Explanation}\par

When we want our learnt representations to be human-understandable, we can tie them to either an \textit{explicitly specified `concept'}  from natural human language or we can learn \textit{`prototypes'} which are \textit{implicitly derived}. Prototypes are semantically relevant visual explanations often represented by the closest training image, parts of an image, or via decoding approaches \cite{liDeepLearningCaseBased2018}. In \textit{concept learning}, one tries to associate known semantic concepts to latent spaces  ~\cite{kohConceptBottleneckModels2020a,kazhdanNowYouSee,fongNet2VecQuantifyingExplaining2018,fangConceptbasedExplanationFinegrained2020,caoConceptLearnersFewShot2021}. 
However, the availability of datasets with annotated concepts or even the prior knowledge of the expected concepts are quite limited. In such cases, \textit{prototype-based learning} using an IBD method provides an alternative to learn global explanations without the need for concept-specific annotations. 

\section{Interpretable Prototypes: Desiderata} 
\label{sec:desiderata}

When we assume \textit{concepts} as something explicitly specified, we basically refer to particular examples that we can recollect from memory (e.g. \textit{bird with red bill}). In case of \textit{prototypes}, they are often average representation over several observed examples \cite{stammerInteractiveDisentanglementLearning2022}.  While the task is to learn independent underlying representations as prototype vectors, a precise visualisation of the prototype vectors in a human-understandable format is still challenging. In literature, prototypes have been interchangeably referred to as representations for a full image or semantically relevant sub-parts of it. In this work, we consider the latter usage for fine-grained interpretability and subsequently chalk out desired properties towards learning interpretable prototypes that are beneficial for human assessment of a model:
\begin{enumerate}
    \item \textit{Human-understandable / interpretable: } The visualisation of the prototype vectors should correspond to a distinct human-relevant entity. Often, due to dataset biases, vague interpretations creep in, like contours of objects or background colours, which often lack in definite explanation.
    \item \textit{Semantically disentangled:} Each prototype should represent distinctly different semantic units that can be associated with common interpretation via humans. 
    \item \textit{Semantically transformation invariant:} All prototypes representing one semantic idea should be uniquely represented irrespective of their variability in scale, translation, or rotation angle across different samples.
    \item \textit{Relevant to the learnt task:} The prototypes learnt should add relevant information towards the task learnt by the given ML model. It can either be the whole semantic entity or distinct sub-parts of it. E.g. for a classification problem for \textit{cars}, the prototypes should be parts which are semantically relevant for identifying a car, like wheels, doors etc., whereas a pedestrian is not related to a car directly.
    
\end{enumerate}
Along with all of the above properties for interpretable prototypes, it is also important to focus on learning the prototypes with \textit{minimum concept-level supervision}. We learn prototypes under the assumption that concept-level supervision is difficult and expensive to get. 
In the rest of this work, we focus our investigation on recent advances in prototype-based learning methods presented in Section \ref{sec:protopnet}.

\section{How do we learn - ``what-looks-like-what?"} 
\label{sec:protopnet}

ProtoPNet \cite{chenThisLooksThat2019a} is an image classifier network that learns representations for relevant sub-parts of an image as prototypes. 

\textbf{(i) Learning prototypes:} They are learnt by appending a prototype layer or a latent space to the feature extractor. The prototypes are class-specific (number of prototypes per class is pre-defined) and learnt by employing a cluster and separation loss on top of the cross-entropy loss, which encourages semantically similar samples to cluster together. 
Since height and width of each prototype is smaller than the feature layer, each prototype represents a patch corresponding to the feature layer in latent space and in turn some prototypical part of the whole image $x$. In this prototype layer,  squared $L^2$ distances between the prototype $p_j$ and all patches of $z$, having similar sizes as $p_j$,  are calculated and inverted to obtain \textit{similarity} scores $S$.  

\textbf{(ii) Visualising prototypes:} The similarity scores together constitute $m$ activation heatmaps $g_{p_j}$ of same spatial size as $z$. They indicate where and how strongly a given prototype is present in $z$ and are reduced to a single similarity score using global max pooling. Based on maximum similarity scores after comparison with all inputs from a prototype-specific class, each prototype is projected onto the nearest $z$. Since the spatial arrangement is preserved in the heatmaps, they can be easily upsampled and overlayed on the full image. The patch corresponding to the maximum similarity score from the heatmap projected upon the input image is thus visualised as a \textit{prototype}.


In this work, we also analyse two successive methods proposing solutions for following shortcomings concerning the aspects of learning and visualising prototypes: 

\textit{(a) Optimising visualising prototypes by addressing the problem of coarse and spatially imprecise upsampling:} By upsampling the low resolution heatmap for most relevant regions of interest for both prototype training and test image, ProtoPNet tries to bring forth the decision ``this relevant prototype from this training image looks like that feature of that test image''. However, the effective receptive field in the original image is much larger. Due to model-agnostic upsampling, the region of interest in the final input image tends to imprecisely cover a lot more than the relevant pixels. To address this problem, \cite{gautamThisLooksMore2021} proposed a method called Prototypical Relevance Propagation (\textbf{PRP}) which builds upon the principles of LRP. It aims to attain more accurate fine-grained model-aware explanations by backpropagating the relevances of the prototypes in ProtoPNet.

\textit{(b) Improved learning of prototypes without a fixed number of prototypes per class:} Authors of ProtoPNet advocated equal representation via a fixed number of prototypes per class leading to a lot of prototypes for further analysis. \cite{nautaNeuralPrototypeTrees2021} propose to incorporate a soft decision tree, called \textbf{ProtoTree}, as a hierarchical model looking into a sequence of decisions through each node prototype to arrive at every test sample. ProtoTree, being IBD by design, allows retraceable decisions mimicking human reasoning while reducing the number of prototypes to only 10\% of ProtoPNet.

\section{Experiments and Discussion }
\label{sec:3}
In this section, using different experiments we perform  analysis of the interpretability of the results from the prototype-based learning methods presented in \ref{sec:protopnet}. We first look at image classification tasks, where we consider real-world datasets and much simplified synthetic dataset in Section \ref{sec:results-real}, \ref{sec:results-synthetic} respectively in light of the \textit{desiderata} in Section \ref{sec:desiderata}. In Section \ref{sec:results-userstudy}, we provide quantitative statistics of our findings in \ref{sec:results-real} based on user-study. Finally, in Section \ref{sec:results-ood}, we present a preliminary application of interpretable prototypes in a real-world OOD detection task. 

\subsection{This Looks Like That? - Analysis on real data}
\label{sec:results-real}

Here, we provide insights on the interpretability of learnt prototypes ProtoPNet, ProtoTree and PRP.
Experimental setups have been kept identical to the respective original implementations.  For PRP, we have reproduced the code as close as possible to mentioned algorithms in \cite{gautamThisLooksMore2021}. Datasets used for fine-grained and generalised image classification are respectively Caltech UCSD Birds-200-2011 ($200$ bird classes) \cite{wah2011caltech} and ImageNet-30 ($30$ distinct classes) \cite{hendrycks2019using}. ProtoPNet uses ImageNet pre-trained VGG19 models. ProtoTree uses ResNet-50 models pre-trained on Naturalist for CUB-200 and ImageNet for ImageNet-30. Following insights are drawn from the entire test data. Let's take a closer look at each method based on the following properties:

\begin{figure*}
    \centering
    \begin{subfigure}{\textwidth}
        \centering
        \includegraphics[width=0.9\textwidth]{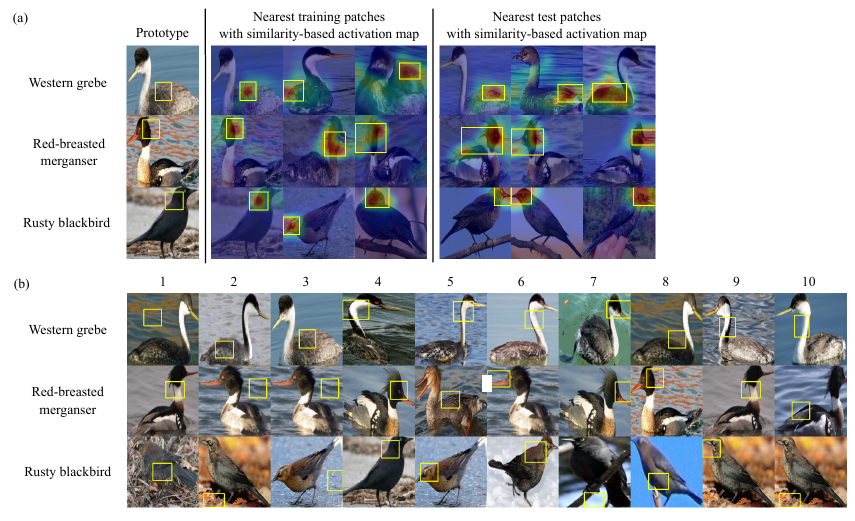}
        \caption*{CUB-200}
    \end{subfigure}
    \vfill
    \begin{subfigure}{\textwidth}
        \centering
        \includegraphics[width=0.9\textwidth]{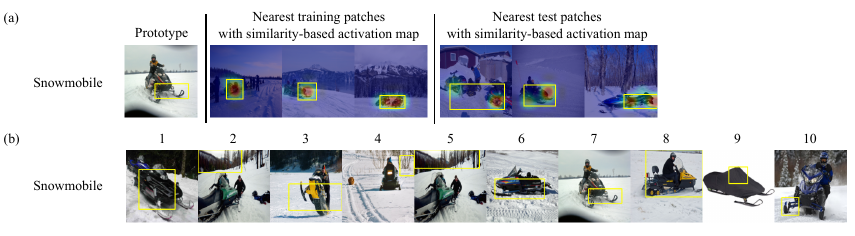}
        \caption*{ImageNet-30}
    \end{subfigure}
    
  \caption{Results for ProtoPNet using CUB-200 (top) and ImageNet-30 (bottom): (a) shows for a prototype from a given class - the nearest training and test patches (yellow box) with similarity score based activation maps, (b) shows all the 10 prototypes (yellow box) learnt for the respective classes in (a).} 
  \label{fig:protop}
\end{figure*}

\subsubsection{Human-Understandable / Interpretable} 
\label{sec:results-real_interpret}
\textit{ProtoPNet:} In Figure \ref{fig:protop}, we present samples from models trained on CUB-200 and ImageNet-30 with $75.9$\% and $97.0$\% test accuracy. In sub-figures (a), we show the $3$ closest train and test images for a given prototype from each class. In a broader sense, the prototypes, given the context where they are located, bring forth the understanding that this patch in the test image \textit{probably} looks like that prototype. Most prototypes can be successfully matched to somewhat similar patches in test images. But the `standalone' prototypes themselves are not so human-interpretable such that they can be distinctly identified as a relevant entity. E.g. for \textit{rusty blackbird}, the prototype shows the neck of the bird, however, from the similarity activation maps for closest test images, the highest similarity varies from eyes, beak to neck region. This shows that the imprecise upsampling of the similarity activation maps often leads to spurious identification of non-relevant parts. Similarly, for \textit{red-breasted merganser}, the probable prototype showing the backside of its head is confused with its beak, head or eyes. For the \textit{snowmobile} class of ImageNet-30, the skis prototype is matched inaccurately to wheels, tracks and even the whole vehicle in test images, thus leading to inconclusive interpretations.

\begin{figure*}
    \centering
    \includegraphics[width=0.75\textwidth]{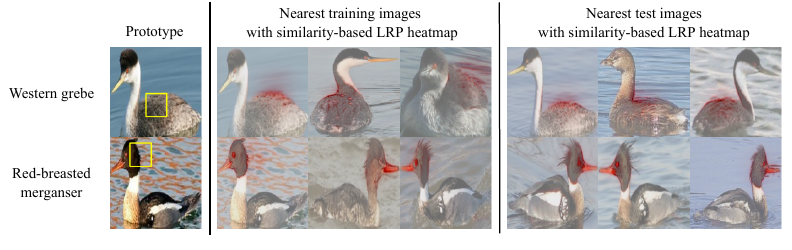}
    \caption{Results using PRP to enhance interpretations of ProtoPNet for corresponding CUB-200 classes. The highlighted regions in red correspond to maximum positive activations corresponding to each prototype.}
  \label{fig:prp}
\end{figure*}

\begin{figure*}
    \centering
    \begin{subfigure}{0.9\textwidth}
        \centering
        \includegraphics[width=\textwidth]{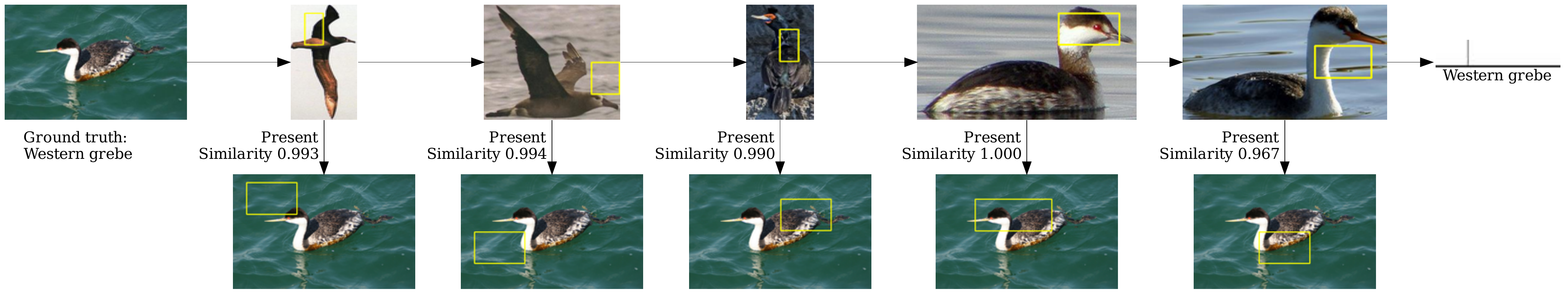}
        \caption*{CUB-200}
    \end{subfigure}
    \vfill
    \begin{subfigure}{0.9\textwidth}
        \centering
        \includegraphics[width=\textwidth]{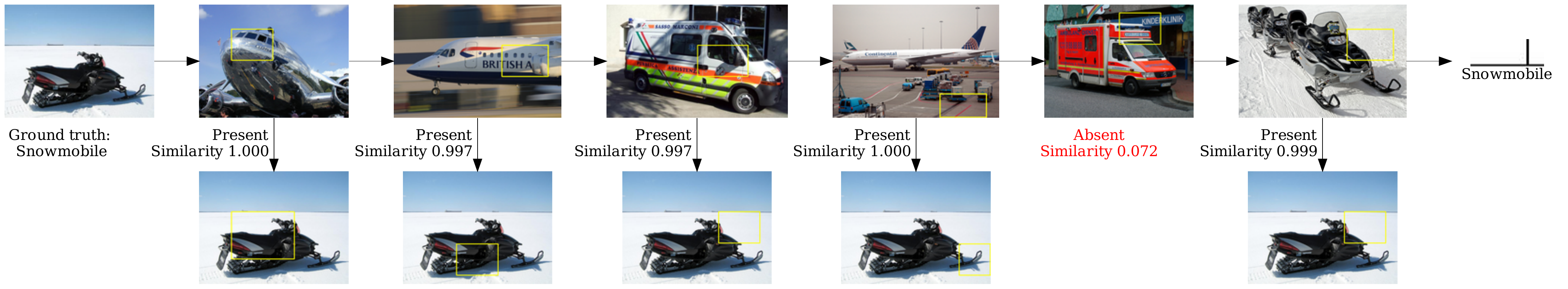}
        \caption*{ImageNet-30}
    \end{subfigure}
     \caption{Results from ProtoTree on CUB-200 (top) and ImageNet-30 (bottom). For each test image, corresponding path taken in the decision tree towards final prediction is shown. The node prototypes at each decision-making stage are shown in yellow boxes in their respective source images along with similarity scores to respective matching parts in test images. Absent node prototypes are marked in red.} 
  \label{fig:tree}
\end{figure*}

\textit{PRP:} 
We compare similar classes in Figures \ref{fig:protop}(a)  and \ref{fig:prp} for CUB-200 to find potential improvement using PRP for the imprecise upsampling mentioned above. For \textit{western grebe}, PRP reassuringly highlights the edges of the upper body as compared to the imprecise body or tail of the bird shown in ProtoPNet. For \textit{red-breasted merganser} however, PRP highlights the beaks, while the prototype looks at the backside of the head, leading to incoherent interpretations. In most samples, PRP tends to focus on the closest edges that might be salient when matched to a prototype. Although these explanations are seemingly more precise compared to ProtoPNet, this does not always enhance the certainty or conviction of the relevant parts for human interpretation.

\textit{ProtoTree:} 
Figure \ref{fig:tree} shows samples from models trained using ProtoTree on CUB-200 and ImageNet-30 with $82.1$\% and $91.8$\% test accuracies. In order to look at the most relevant prototypes used for deciding on a class, we chose classes from the rightmost branch of the tree to ensure the traversed node prototypes were `present' in most cases in the decision path for these classes, namely \textit{western grebe}  and \textit{snowmobile} from CUB-200 and ImageNet-30. As pointed out by the authors, since the prototypes themselves are not mapped to any particular class, the first prototypes in the given path are barely relevant for a given class or the indicated matching parts. Prototype 1 in \textit{western grebe} is hardly understandable, similarly the third prototype looks at the black body of a bird but individually the prototype is difficult to comprehend.
Thus, the prototypes themselves are not entities easily understandable to human-eye particularly, even more  for fine-grained image datasets like CUB which require expert knowledge.

\subsubsection{Semantically Disentangled}

Since ProtoPNet and ProtoTree learn the prototypes differently, here we analyse whether they successfully learn distinct prototypes corresponding to distinct semantically relevant parts.

\textit{ProtoPNet:} In line with the ProtoPNet implementation, $10$ prototypes are learnt per class as shown in Figure \ref{fig:protop}(b). We observe that the learnt prototypes are often redundant, i.e. similar prototypes or prototypes looking at similar parts. E.g. for the \textit{western grebe} class- prototypes $4, 5, 6, 9,$ and $10$ show neck parts, similarly \textit{red-breasted merganser} shows repeated neck ($1, 9$)  and background prototypes ($2, 3$), even from the same training image. Most repeated prototypes do not add a different perspective in terms of looking at different details of a semantic part, e.g. there are multiple ski parts (prototypes $3, 7,$ and $10$) in \textit{snowmobile}. Overall, prototypes need to be more distinct and diverse to ensure complete mutually exclusive representation of the entire class. The redundancy could be due to too many pre-determined prototypes for each class. Thus, we need methods better suited to fine-tune the optimal number of prototypes to the given dataset and respective classes.

\textit{ProtoTree:} In Figure \ref{fig:tree}, although much lesser number of prototypes are learnt ($10\%$ of ProtoPNet) avoiding redundancy and fewer background prototypes, most of them are not class-specific thus quite semantically disentangled over the whole dataset. The prototypes are so diverse that it is difficult to semantically correlate to the matching parts of class-specific test images, thus providing limited interpretations towards learning semantics of any particular class. 


\begin{figure*}
    \centering
    \includegraphics[width=0.7\textwidth]{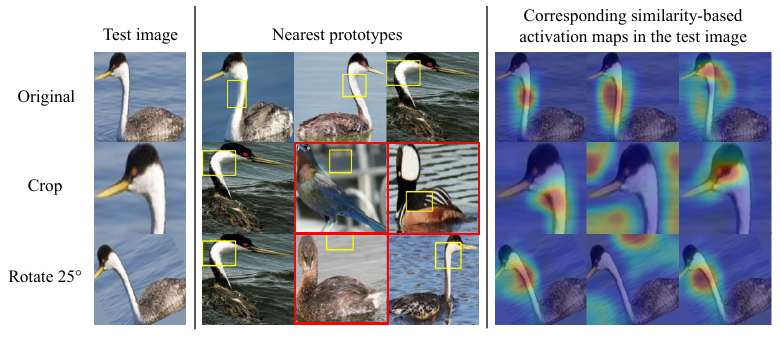}
    \caption{ProtoPNet results corresponding to transformations (crop, rotate $25^{\circ}$) for a sample test image (\textit{western grebe}) showing the nearest $3$ prototypes and their respective region of activations in the given test image. Yellow boxes show the prototypes and images in red box show prototypes taken from a different class than \textit{western grebe}.} 
    \label{fig:transfroms}
\end{figure*}

\subsubsection{Semantically Transformation Invariant}
Given a prototype, say - the head of a bird, it is essential that our methods learn representations for these prototypes irrespective of variations that appear for this particular entity across the entire dataset, i.e. prototypes should be transformation invariant. Since these methods use $L^2$ similarity in the feature space for matching relevant test image patches, it is important to inspect this property using transformed (rotated and cropped) test samples.

\textit{ProtoPNet:} In Figure \ref{fig:transfroms}, we show how well prototypes are recognised given a transformed version of the test image. We show the top $3$ prototypes given a test image and respective patches they activate in terms of maximum similarity score. We see that given a cropped head of \textit{western grebe}, $2$  out of the top $3$  closest prototypes belong to different classes. Similarly, given a rotated version of this same test image by $25^{\circ}$, one of the closest prototypes is a background prototype from a different class. 

\textit{ProtoTree:} Repeating above experiments with ProtoTree, we note that the node prototypes and the path in the tree for the transformed test images did not change, indicating ProtoTree to be more robust to image transformations than ProtoPNet. Thus, we do not show these results to save space and avoid redundancy.
Since ProtoTree and ProtoPNet use a different set of augmentations during training, we also trained a ProtoPNet with the augmentations applied in ProtoTree (different crops etc.), but this did not improve the performance of ProtoPNet against transformations. Possibly, ProtoTree learns much fewer prototypes and thus larger semantic distances between prototypes, making it robust to smaller semantic transformations. However, this needs further investigation.

\subsubsection{Relevant to the Learnt Task}
\label{sec:results-real_relevant}

Considering the classification task as a \textit{\lq Guess who?'} game where by looking at the learnt prototypes, can we guess the collective class they belong to? 

\textit{ProtoPNet:} As observed earlier, the $10$ prototypes for each class shown in  \ref{fig:protop}(b) are often redundant and do not always represent all the distinct parts of their respective classes. Nonetheless, some prototypes do provide hints towards the respective classes to make an informed guess like the white neck prototype for \textit{western grebe} hints at a bird with a white neck, or the prototypes showing a black neck, head or wings for \textit{rusty blackbird} indicate at least a black bird. However, whether they sufficiently represent their respective class remains doubtful, particularly for fine-grained dataset. For more generalised datasets like ImageNet-30, the prototypes themselves are quite diverse and easier to recognise, often including the entire object in question, e.g. \textit{snowmobile} prototypes ($1, 8$) showing the whole vehicle. There are several instances of background prototypes ($2, 4, 5$ in \textit{snowmobile}) which might provide some context to recognise a given class, but in general add to redundancy. Lastly, it is up to the interpreter to make sense out of these disjoint bits of information.

\textit{ProtoTree:} ProtoTree itself does not provide class-specific prototypes, thus all the node prototypes in a given path which are marked as `present' (see Figure \ref{fig:tree}) are often not relevant directly to the class in question except for the last few prototypes, e.g. the bird's eye or the neck for \textit{western grebe}. Looking at the prototypes of the \textit{snowmobile} class, one would almost guess it as an airplane class except for the last prototype with snow.
Thus, while prototypes from ProtoPNet provide some reliable hints as compared to ProtoTree, both methods perform insufficiently in a \textit{`Guess who?'} game.

\subsection{Quantitative results based on User Study}
\label{sec:results-userstudy}

\begin{figure*}
    \centering
    \begin{subfigure}{0.35\textwidth}
        \centering
        \includegraphics[width=\textwidth, height= 4cm]{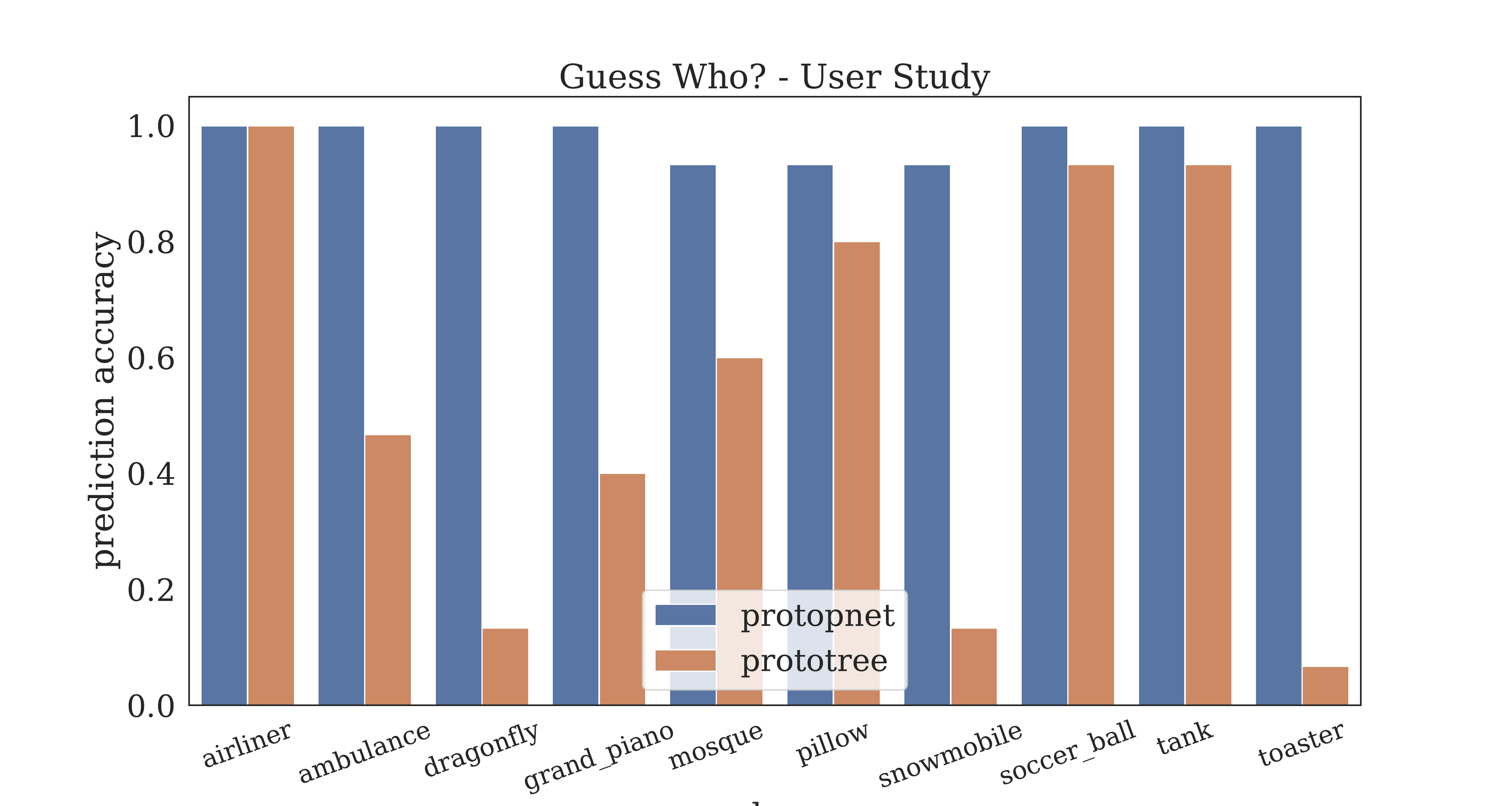}
        \caption*{Class prediction accuracy}
    \end{subfigure}
    \hfill
    \begin{subfigure}{0.3\textwidth}
        \centering
        \includegraphics[width=\textwidth, height=2.5cm]{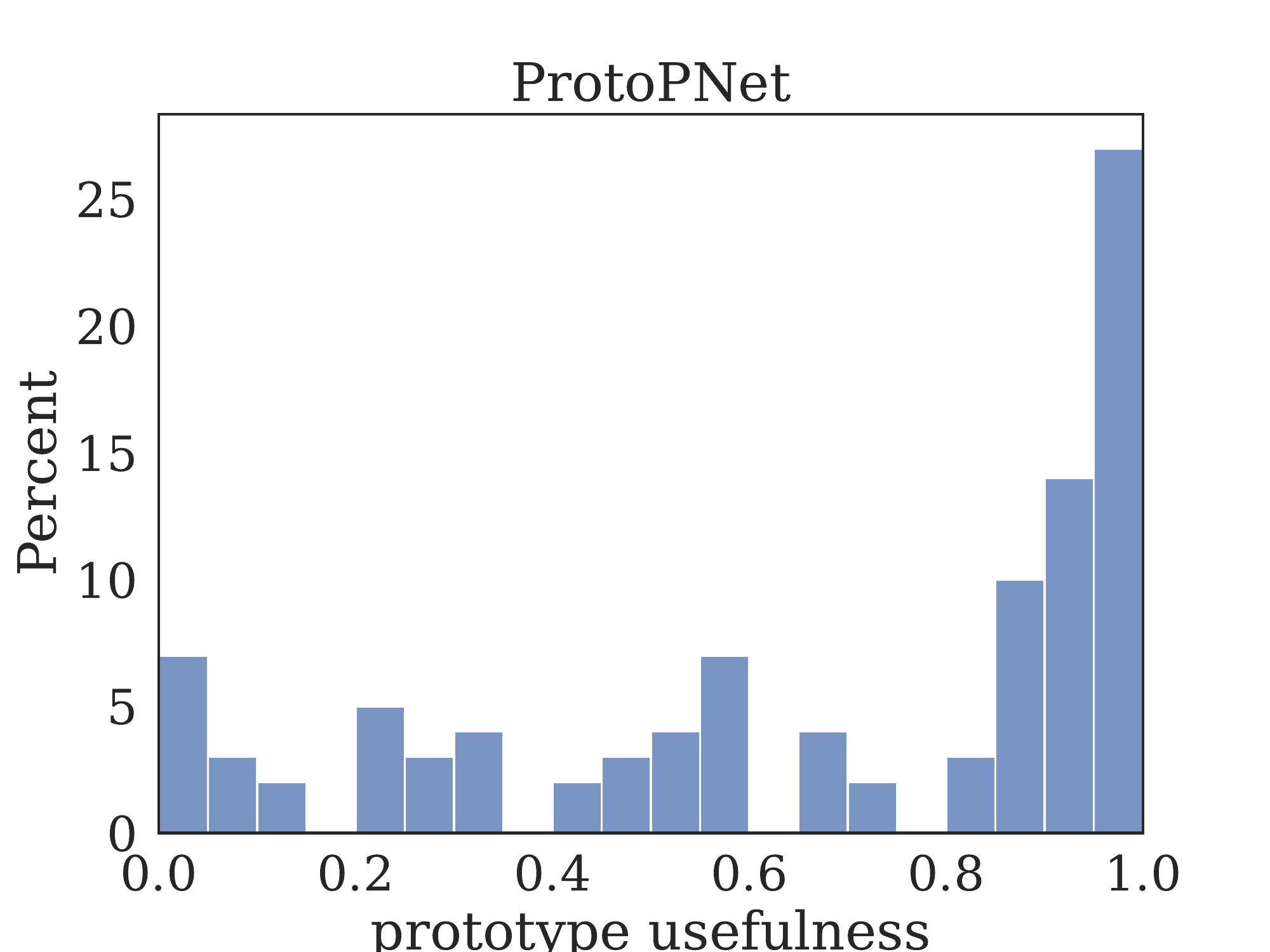}
        \includegraphics[width=\textwidth, height=2.5cm]{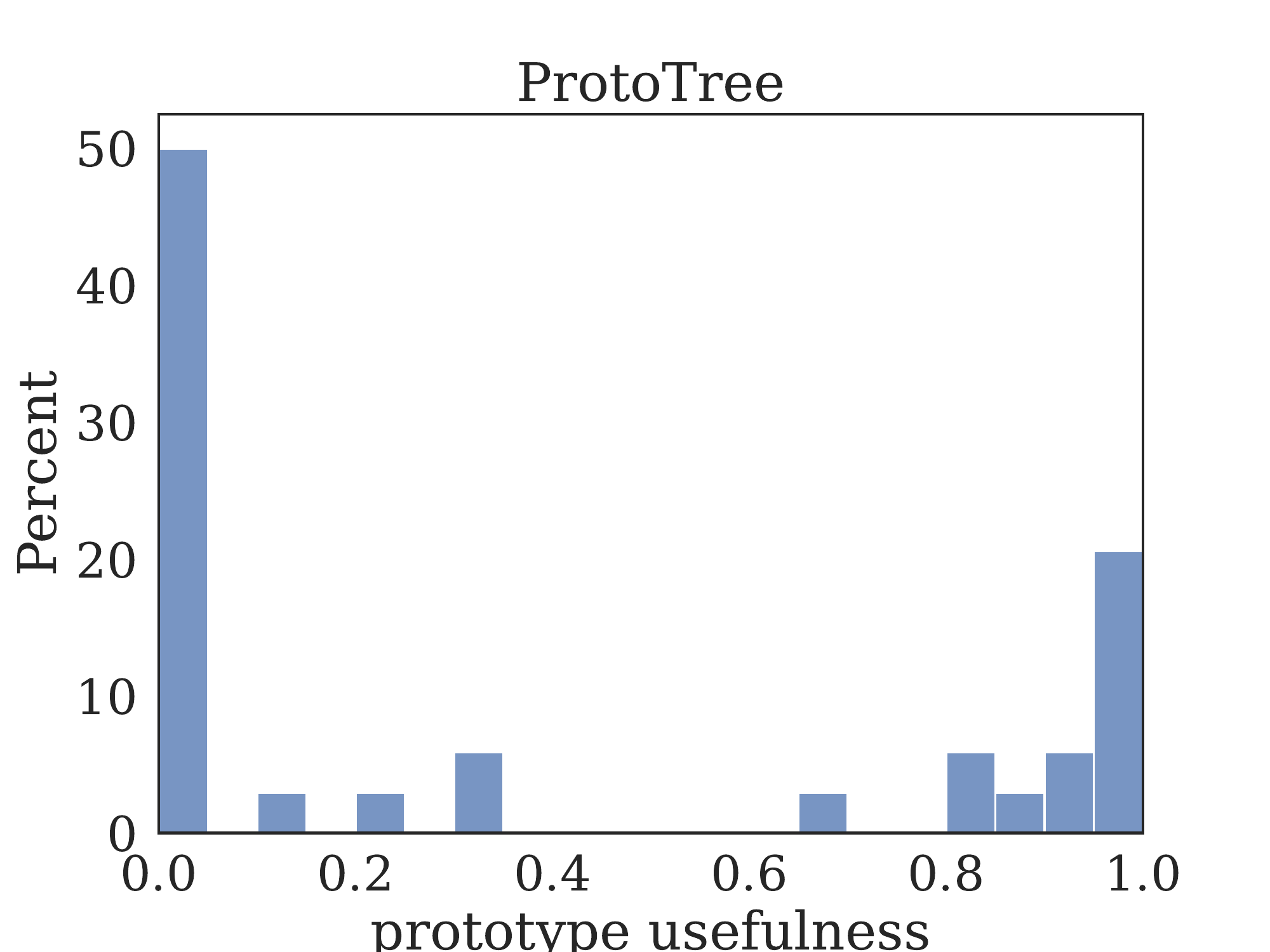}
        
        \caption*{Prototype usefulness}
    \end{subfigure}
    \hfill
    \begin{subfigure}{0.3\textwidth}
        \centering
        \includegraphics[width=\textwidth, height=2.5cm]{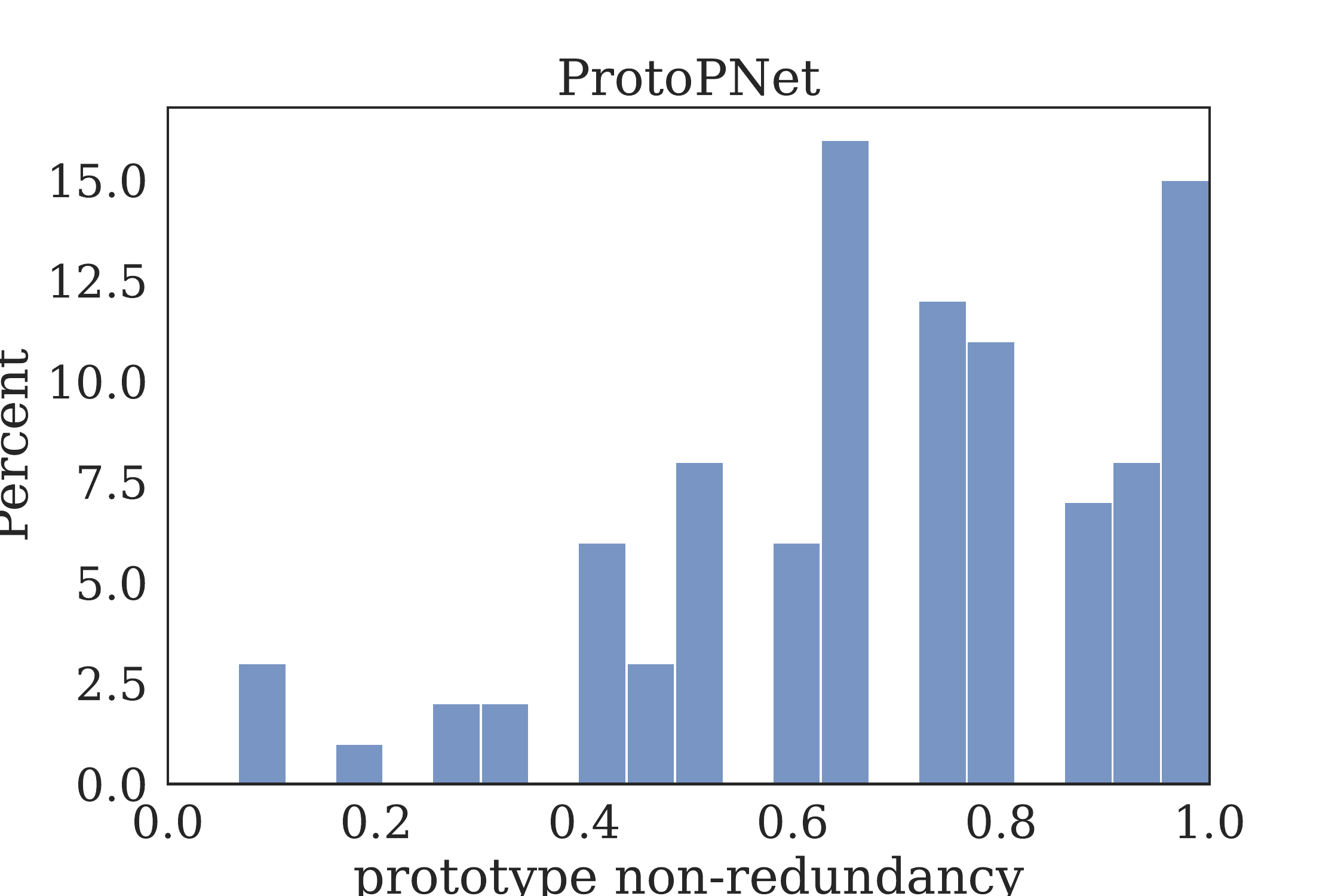}
        \includegraphics[width=\textwidth, height=2.5cm]{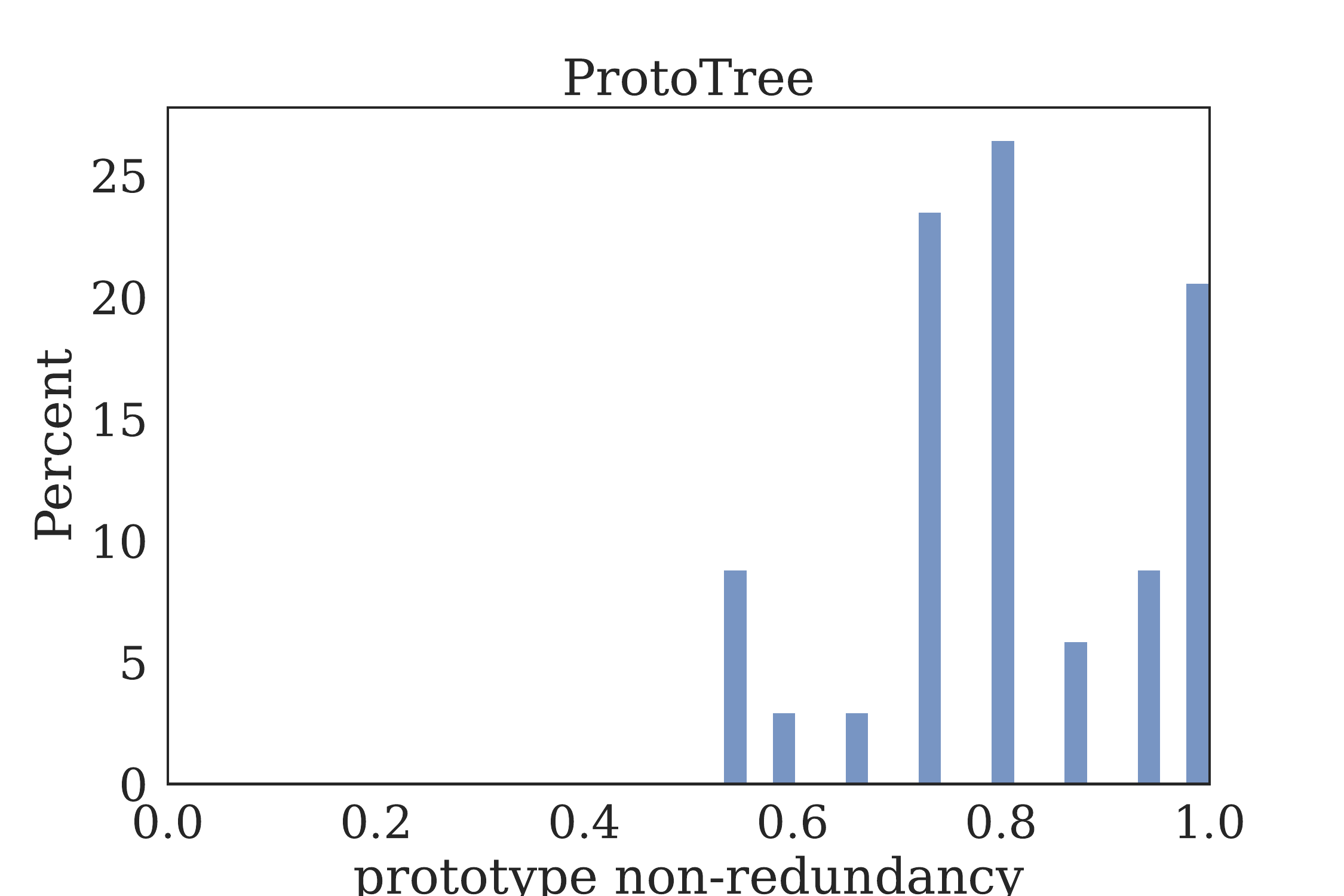}
        \caption*{Prototype non-redundancy}
    \end{subfigure}
    
  \caption{Quantitative results from user study conducted on prototypes from ImageNet-30 classes} 
  \label{fig:user_study}
\end{figure*}

To further validate our observations in Section \ref{sec:results-real}, we collected statistics based on human assessment of prototypes of natural images ($10$ classes of ImageNet-30) to avoid the need of expert knowledge for fine-grained datasets. The study comprised of $2$ experiments with $15$ users: (1) given prototypes, users were required to identify the class  (\textit{`Guess who?'} game) (2) given each prototype and respective class, users were asked to determine whether they were `useful' and `redundant'. Detailed experiments are given in Appendix. 
Figure \ref{fig:user_study} summarises the results from all experiments, which could be interpreted as follows based on the questions asked to the users:

 \textit{`Guess Who?'} - Average class prediction accuracy for ProtoPNet was much higher ($98 \%$) as compared to ProtoTree ($55 \%$). This is as expected as ProtoPNet prototypes could easily be guessed from given ten classes as common sub-parts of natural images. However, for ProtoTree, many of the prototypes do not belong to the class and the initial prototypes on a given branch of a tree are often general and irrelevant to the class.Thus, these prototypes were often not semantically relevant or human understandable and thus, difficult to identify leading to poor prediction accuracy for ProtoTree.
    
\textit{Prototype usefulness} - Only $27 \%$ prototypes of ProtoPNet and $20 \%$ of ProtoTree, were found totally useful ($100 \%$) for identifying the class. This leaves a lot of future scope to generate semantically relevant and yet diverse prototypes that represents the given class sufficiently for confident human interpretation.
    
\textit{Prototype non-redundancy} - Since we had observed a lot of redundant or repeating prototypes in cases of fine grained datasets like CUB-200, hence this experiment was conducted. However, as we noted earlier that for natural images, the prototypes are much more diverse and non-redundant. Hence, for this experiment, it leaves an ambiguity whether the prototypes were actually non-repeating or whether they were found irrelevant/meaningless and thus marked as non-redundant ($15 \%$ for ProtoPNet and $20.6 \%$ for ProtoTree).
    
Using the above statistics, we could further confirm the need for better methods that can generate truly human-interpretable prototypes that are semantically relevant, disentangled as well as sufficient to identify a given class.

\begin{figure}[!htb]
    \centering
    \begin{subfigure}{0.8\linewidth} 
        \centering
        \includegraphics[width=\textwidth, height=4cm]{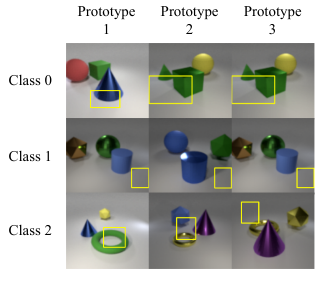}
        \caption{ProtoPNet}
    \end{subfigure}
    \vfill
    \begin{subfigure}{0.8\linewidth} 
        \centering
        \includegraphics[width=\textwidth, height=4cm]{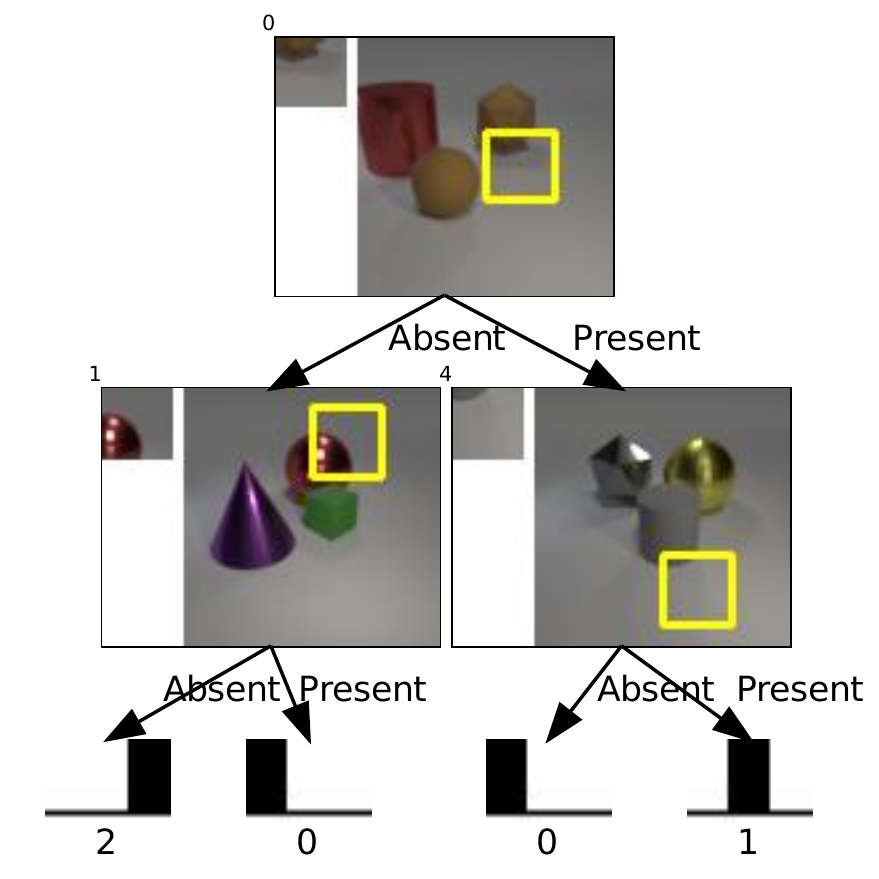}
        
        \caption{ProtoTree}
    \end{subfigure}
     \caption{Results using synthetic 3D-Shapes dataset ($V_1$) showing in (a) prototypes from each class using ProtoPNet and (b) the node prototypes along with decision tree (depth=2) using ProtoTree. Generated dataset $V_1$ consists of $3$ classes with $3$ non-exclusive shapes- class $0$ (\textit{cube, sphere, cone}), class $1$ (\textit{sphere, cylinder, icosphere}) and class $2$ (\textit{cone, torus, icosphere}). Yellow boxes show the prototypes.} 
  \label{fig:3d_shapes}
\end{figure}

\subsection{Analysis on synthetic data}
\label{sec:results-synthetic}

In previous sections, we observed that it is difficult to obtain human-interpretable prototypes with the wide range of complexities associated with real-world datasets, like the optimum number of prototypes required to define a class, varying semantics, cluttered background, overlapping concepts, etc. Thus, we created synthetic datasets (\textit{3D-Shapes}) in a controlled setting where each shape can be related a prototypical concept and re-evaluated the performance of above methods. The 3D-Shapes datasets consist of combinations of rendered 3D shapes in varying arrangements\cite{Johnson_2017_CVPR}.

\textit{Dataset with overlapping concepts ($V_1$):}  This dataset consists of $3$ classes with $3$ non-exclusive shapes each, resembling the original fine-grained classification setting. Exemplary results for ProtoPNet and ProtoTree are given in Figure \ref{fig:3d_shapes}. Ideally, each shape in a class should correspond to a prototype. Like in Section \ref{sec:results-real} for \textit{ProtoPNet}, we observe redundant repeating prototypes even in this simplified setting. Also all the prototypes of one class focus on the background, however they still contribute to a high test accuracy. The learnt prototypes are not semantically relevant in a way humans would classify 3D shapes. Often they focus on semantically mixed patches like parts of both cube and cone in class $0$. Similarly, in \textit{ProtoTree} the decision paths using the highlighted prototypes do not follow human logic, as e.g. the path to class $2$ is reached via $2$ absent prototypes (partial \textit{icosphere}, \textit{sphere}), whereas the \textit{icosphere} should belong to class $2$. Instead it arrives at class $2$ by eliminating images from class $1$ and $0$.Thus, the prototypes are not semantically relevant for classification nor matching to the class.

\textit{Dataset with non-overlapping concepts ($V_2$):} This dataset is designed to be even simpler with $3$ classes composed of $2$ shapes each which are mutually exclusive with other classes. For each class, we expect that two distinct prototypes corresponding to each shape should be learnt. For \textit{ProtoPNet}, we observe that both prototypes for each class are similar (redundant) which is relevant to the classification task as there is no incentive to learn the other shape; however, for practical use-cases we expect it to learn distinct and diverse prototypes. With limited number of concepts, no more background prototypes are observed, though the learnt prototypes often focus on mixed patches and are not semantically human-understandable. In \textit{ProtoTree}, the decision tree mostly follows a logical structure es expected by a human. However, some prototypes are still not human-understandable as they do not match to the corresponding indicated parts, for e.g. the prototype with partial edges of the \textit{cube} somehow finds the \textit{sphere} in the test image as the corresponding matching part. 

Details regarding above datasets and experiments are given in Appendix. Ideally, recognising a class based on interpretable prototypes as prior evidence would help us make more informed classification decisions particularly for safety-critical use-cases. 

\subsection{Application for Real-World Tasks}
\label{sec:results-ood}
Taking cue from the above mentioned properties, where we assume ideally prototypes are human-interpretable and semantically disentangled, one potential real-world application could be to distinguish OOD samples from samples belonging to ID classes. It is assumed that OOD samples would have very different prototypes as compared to ID. 
\begin{figure}
    \centering
    \includegraphics[width=0.48\linewidth]{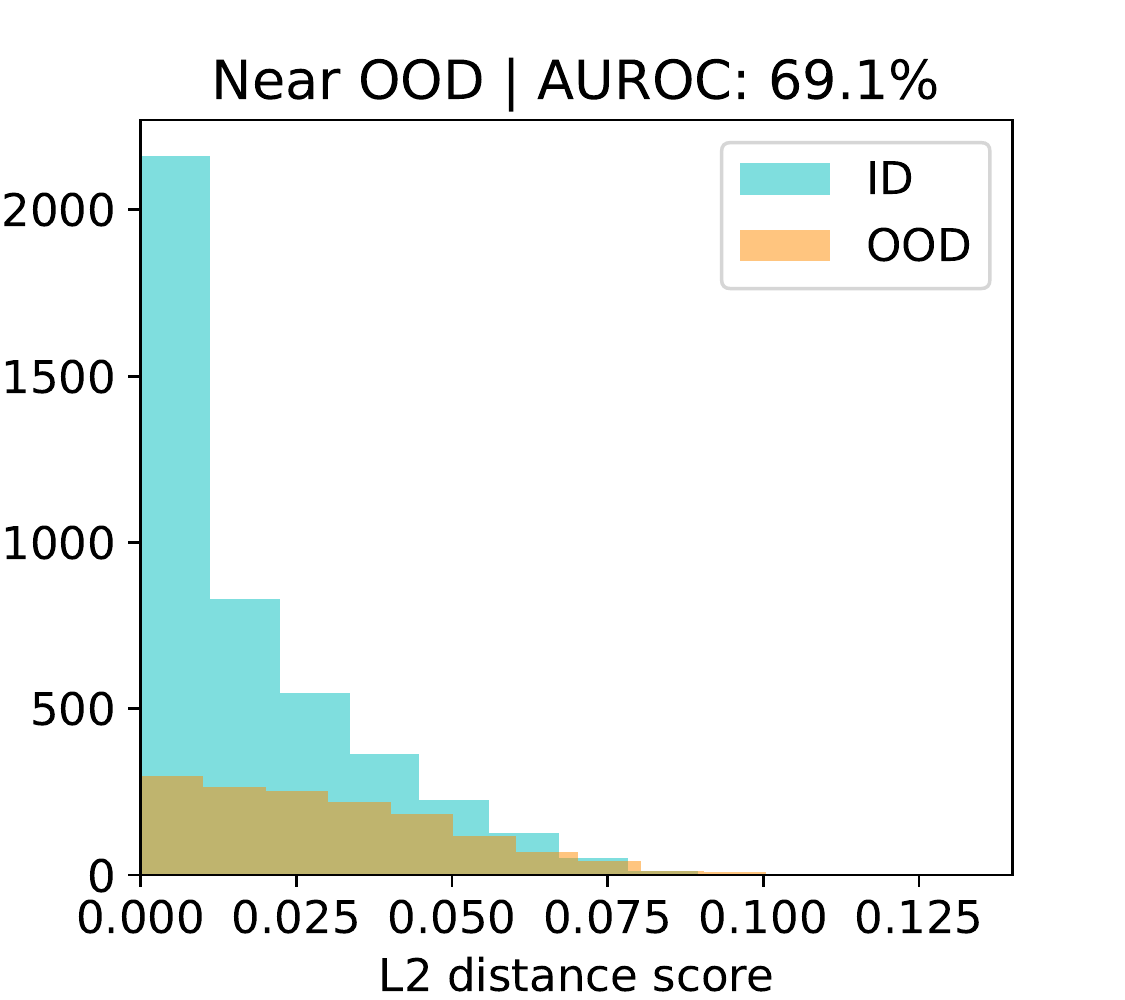}
    \includegraphics[width=0.48\linewidth]{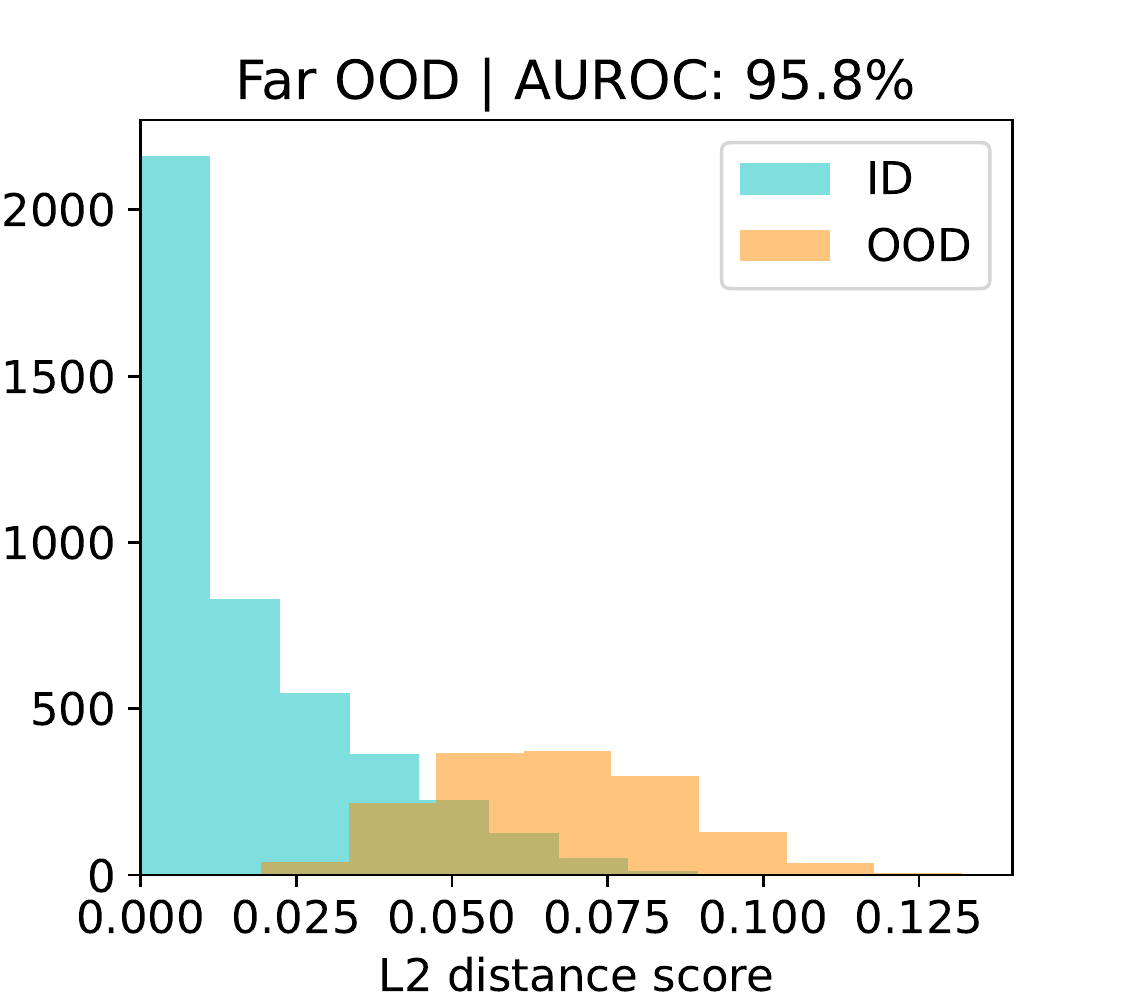}    
    \caption{Histogram showing distribution of $L^2$ distances to closest prototypes for Near vs Far OOD samples for a model trained on first $150$ classes of CUB-200 dataset.} 
    \label{fig:ood}
\end{figure}
A simple approach would be to distinguish test ID and OOD samples based on their $L^2$ distances to the nearest prototypes. We train a ProtoPNet on $150$ (out of $200$) bird classes from the CUB-200 dataset as ID training data. The remaining $50$ bird classes serve as `Near OOD' data. As the prototypes from this OOD data are still from the birds dataset, they are expected to be semantically similar to ID prototypes. SVHN \cite{netzer2011reading}, a dataset consisting of housing numbers, is taken as `Far OOD' data. The Area Under ROC curve (AUROC) provides an evaluation metric for OOD detection and the results are shown in Figure \ref{fig:ood}. We observe a lot of OOD samples having closely overlapping $L^2$ distances in `Near OOD' setting, which is concurrent with the fact that OOD prototypes are very similar to ID prototypes and thus performs poorly in terms of AUROC ($69.1\%$). In `Far OOD' setting, SVHN samples are distinctly separated in terms of $L^2$ distances from training prototypes, which leads to an AUROC of $95.8\%$ in terms of OOD detection. We remark that this approach might not be an absolute representative of the separation of ID and OOD prototypes in the feature space. 

\section{Conclusions and Future Work}
\label{sec:conclusion}
In this work, we have assessed the interpretability of the prototypes learnt from various prototype-based IBD methods in terms of visual relevance to humans. To that end, we first defined a set of desired properties of the prototypes as a basis for our analysis of three different approaches: ProtoPNet, ProtoTree and Prototypical Relevance Propagation (PRP). We found ProtoPNet generates somewhat relevant prototypes but suffers from a lot of redundancy and a lack of semantically distinct prototypes. ProtoTree produces semantically diverse prototypes which are less redundant but mostly not relevant. PRP addresses the imprecise upsampling of ProtoPNet but does not conclusively contribute to better interpretability. Overall, standalone prototypes individually (without matching location context) are mostly not human-interpretable and there is still a long way to go.    

Potential future work should focus more on improving the quality of the learnt prototypes in terms of valuable human-understandable interpretations as well as explore techniques to diversify the prototypes to avoid redundancy. One way to improve quality of explanations as well as improve trustworthiness in high-stake decisions would be to utilise human feedback during the learning phase to identify useful prototypes as a potential next step. This could also strengthen the need to demonstrate and validate which properties are actually required for interpretability and for effective internal assessment of models. As observed already, given relevant prototypes, OOD detection could largely benefit from interpretable prototypes which calls for finding better techniques, particularly in `Near OOD' regime. 

\bibliography{aaai23}

\end{document}